\documentclass[runningheads]{llncs}
\usepackage[T1]{fontenc}
\usepackage{fontawesome}
\usepackage{marvosym}
\usepackage{booktabs}     
\usepackage{tabularx}      
\usepackage{makecell}     
\usepackage{amsmath}
\usepackage{amssymb}     
\usepackage{xcolor} 
\usepackage{graphicx,verbatim}
\usepackage{hyperref}

\usepackage{multirow} 

\usepackage{stackengine}
\usepackage{array}
\usepackage{arydshln}
\usepackage{pifont}
\usepackage{threeparttable}

\newcommand{\para}[1]{\vspace{.05in}\noindent\textbf{#1}}

\newcommand{\meanstd}[2]{%
    \makecell[c]{$#1$\\\fontsize{8pt}{8pt}$\pm #2$}%
}
\begin{document}
\title{HyperPath: Knowledge-Guided Hyperbolic Semantic Hierarchy Modeling for WSI Analysis}
\titlerunning{HyperPath}
%
%
\author{Peixiang Huang\inst{1} \and
Yanyan Huang\inst{1} \and
Weiqin Zhao\inst{1} \and\\
Junjun He\inst{2,3} \and 
Lequan Yu\inst{1}\textsuperscript{\,(\faEnvelopeO)} }
\authorrunning{P. Huang et al.}
%
\institute{
  \textsuperscript{1\ }The University of Hong Kong, Hong Kong SAR, China \\
  \email{\texttt{\{paxson\_huang, yanyanh, wqzhao98\}@connect.hku.hk, lqyu@hku.hk}}\\
  \textsuperscript{2\ }Shanghai Artificial Intelligence Laboratory, Shanghai, China \\
  \textsuperscript{3\ }Shanghai Innovation Institute, Shanghai, China \\
  \email{hejunjun@sjtu.edu.cn}}
\maketitle              
\begin{abstract}
Pathology is essential for cancer diagnosis, with multiple instance learning (MIL) widely used for whole slide image (WSI) analysis. WSIs exhibit a natural hierarchy—patches, regions, and slides—with distinct semantic associations. While some methods attempt to leverage this hierarchy for improved representation, they predominantly rely on Euclidean embeddings, which struggle to fully capture semantic hierarchies.
To address this limitation, we propose \textbf{HyperPath}, a novel method that integrates knowledge from textual descriptions to guide the modeling of semantic hierarchies of WSIs in hyperbolic space, thereby enhancing WSI classification. 
Our approach adapts both visual and textual features extracted by pathology vision-language foundation models to the hyperbolic space. We design an Angular Modality Alignment Loss to ensure robust cross-modal alignment, while a Semantic Hierarchy Consistency Loss further refines feature hierarchies through entailment and contradiction relationships and thus enhance semantic coherence.
The classification is performed with geodesic distance, which measures the similarity between entities in the hyperbolic semantic hierarchy. This eliminates the need for linear classifiers and enables a geometry-aware approach to WSI analysis.
Extensive experiments show that our method achieves superior performance across tasks compared to existing methods, highlighting the potential of hyperbolic embeddings for WSI analysis. The source code is available at \href{https://github.com/HKU-MedAI/HyperPath }{https://github.com/HKU-MedAI/HyperPath}.

\end{abstract}
\section{Introduction}
Pathology is the gold standard for cancer diagnosis, and whole slide image (WSI) analysis is a key component of computational pathology, advancing cancer diagnosis and prognosis through machine learning~\cite{comprehensivereview}. However, due to the large size and complex patterns of WSIs, pixel-level annotations are impractical.
Multiple Instance Learning (MIL)~\cite{mil} addresses this by operating on bags of image patches without exhaustive labeling, enabling slide-level representation learning for downstream tasks. Some attention-based MIL methods~\cite{abmil,clam,acmil} leverage aggregation operators to combine patch-level information, providing interpretable and effective representations. TransMIL~\cite{transmil} incorporates Transformers to aggregate morphological and spatial features efficiently, while DTFD-MIL~\cite{dtfdmil} introduces pseudo-bags to address challenges posed by small sample sizes. 

\begin{figure}[t!]
\centering
\includegraphics[width=\textwidth]{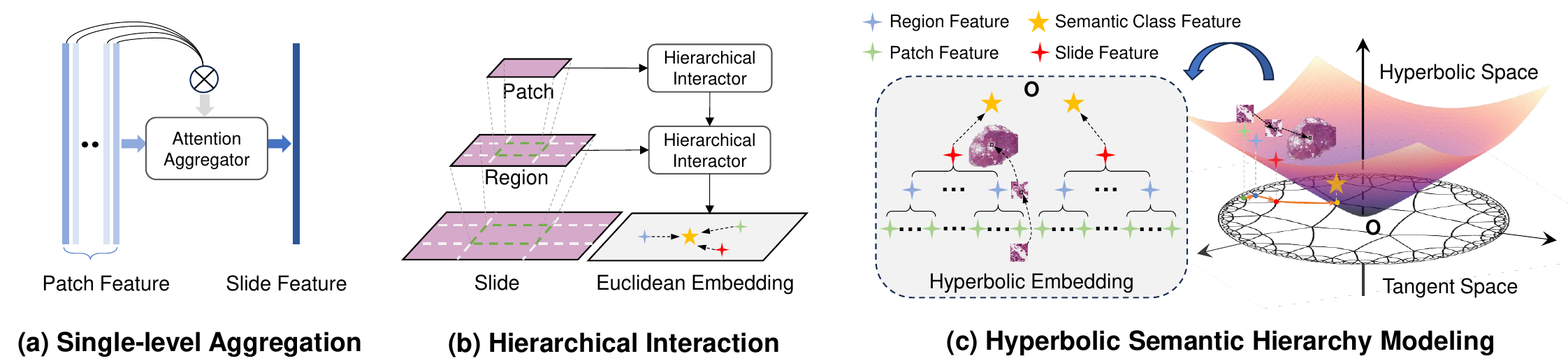}
\caption{Comparison of different representation learning approaches for WSI.} 
\label{fig:teaser}
\end{figure}

Despite these advantages, simple single-level feature aggregation often fails to explicitly model the hierarchical structure of slides. Hierarchical modeling is essential as it captures both local details and global context by representing complex relationships across hierarchical levels. To address this, methods like~\cite{hipt,higt,conslide} extract multi-scale features to model spatial hierarchies. However, these may not fully preserve intrinsic semantic hierarchies.
This limitation has motivated exploration into hyperbolic modeling, a paradigm well-suited for hierarchical structures. Recent studies~\cite{meru,hyperzero,compositional_entailment,acceptgap,vlhyperamazon,learningvisualhierarchy,hypercd} have demonstrated its effectiveness, especially in capturing visual and textual hierarchical relationships.  
In WSIs, hyperbolic modeling can organize levels (patch-region-slide) to align with semantic and hierarchical structures, as more intuitively shown in Fig.~\ref{fig:teaser}.

In this paper, we introduce \textbf{HyperPath}, a novel method that leverages textual concept knowledge to model hierarchical semantic relationships in WSIs within hyperbolic space, improving classification performance. 
Building on the success of foundation models in pathology \cite{uni,cate,conch,virchow,chief}, we employ CONCH~\cite{conch}
for feature extraction. Specifically, our framework encodes image patches and class description prompts into hyperbolic space, where hierarchical aggregation is performed to extract region- and slide-level features that reflect the intrinsic structure of WSIs. By leveraging geodesic distances between slide representations and semantic class features, our method achieves robust WSI classification without relying on linear classifiers.
To enhance the semantic hierarchy in hyperbolic space, we propose two key loss functions: \textbf{Angular Modality Alignment Loss} and \textbf{Semantic Hierarchy Consistency Loss}. The former minimizes cross-modality discrepancies, ensuring effective and robust alignment across hierarchical levels, while the latter enforces semantic coherence within and across modalities, addressing contradictions and promoting entailments. Extensive experiments on four TCGA tasks demonstrate the effectiveness of our approach and its ability to learn semantic hierarchies in WSIs.

\section{Methodology}

\subsection{Preliminaries}

\textbf{Hyperbolic Space.} Hyperbolic geometry exhibits exponential space expansion, allowing it to naturally represent hierarchical structures with efficient scaling, so that it can embed complex relationships without excessive distortion~\cite{hyperimageemb,learnstructure,hyperlearningrevisit}.
Following~\cite{meru}, we choose the Lorentz model to present the $k$-dimensional hyperbolic space with curvature $-\rho$, denoted by $\mathbb{H}_{\rho}^k$, due to its numerical stability and efficiency. It can be described by $(k+1)$-dimensional Euclidean space $\mathbb{R}^{k+1}$, where for every vector $\mathbf{u} \in \mathbb{R}^{k+1}$, the first dimension corresponds to the time component $\mathbf{u}_{t}\in\mathbb{R}$, the remaining dimensions represent the space component $\mathbf{u}_{s}\in\mathbb{R}^{k}$, satisfying $\mathbf{u}_{t} = \sqrt{1/\rho+\|\mathbf{u}_{s}\|_\mathbb{E}^2}$, where $\|\cdot\|_\mathbb{E}$ is the Euclidean norm.
Let the Euclidean and Lorentzian inner product be denoted as $\langle \cdot, \cdot \rangle_\mathbb{E}$ and $\langle \cdot, \cdot \rangle_\mathbb{H}$ respectively, they satisfy the following equation $\langle \mathbf{u}, \mathbf{v} \rangle_\mathbb{H} = \langle \mathbf{u}_s, \mathbf{v}_s \rangle_\mathbb{E}-\mathbf{u}_t \mathbf{v}_t $. Thus the hyperbolic space can be defined as $\mathbb{H}_{\rho}^k = \{ \mathbf{u} \in \mathbb{R}^{k+1} : \langle \mathbf{u}, \mathbf{u} \rangle_\mathbb{H} = -1/\rho, \, \rho > 0 \}$ and the induced Lorentzian norm can be expressed as $\|\mathbf{u}\|_\mathbb{H}=\sqrt{\lvert\langle \mathbf{u}, \mathbf{u} \rangle_\mathbb{H}\rvert}$.

\para{Tangent Space.} 
The tangent space is an orthogonal Euclidean space linked to each point in hyperbolic space, allowing projections that preserve hyperbolic geometry. The origin $\mathbf{O}$ is commonly used as the reference due to its symmetry and simplicity. The transformation $\mathcal{T}_{\mathbb{E}_{\mathbb{T}_{\mathbf{o}}}\rightarrow\mathbb{H}}(\mathbf{x})$ from tangent space $\mathbb{E}_{\mathbb{T}_{\mathbf{o}}}$ to hyperbolic space $\mathbb{H}$ is given by $\mathbf{x}\sinh(\sqrt{\rho}\|\mathbf{x}\|_{\mathbb{E}})/(\sqrt{\rho}\|\mathbf{x}\|_{\mathbb{E}})$.

\begin{figure}[t]
\centering
\includegraphics[width=0.9\textwidth]{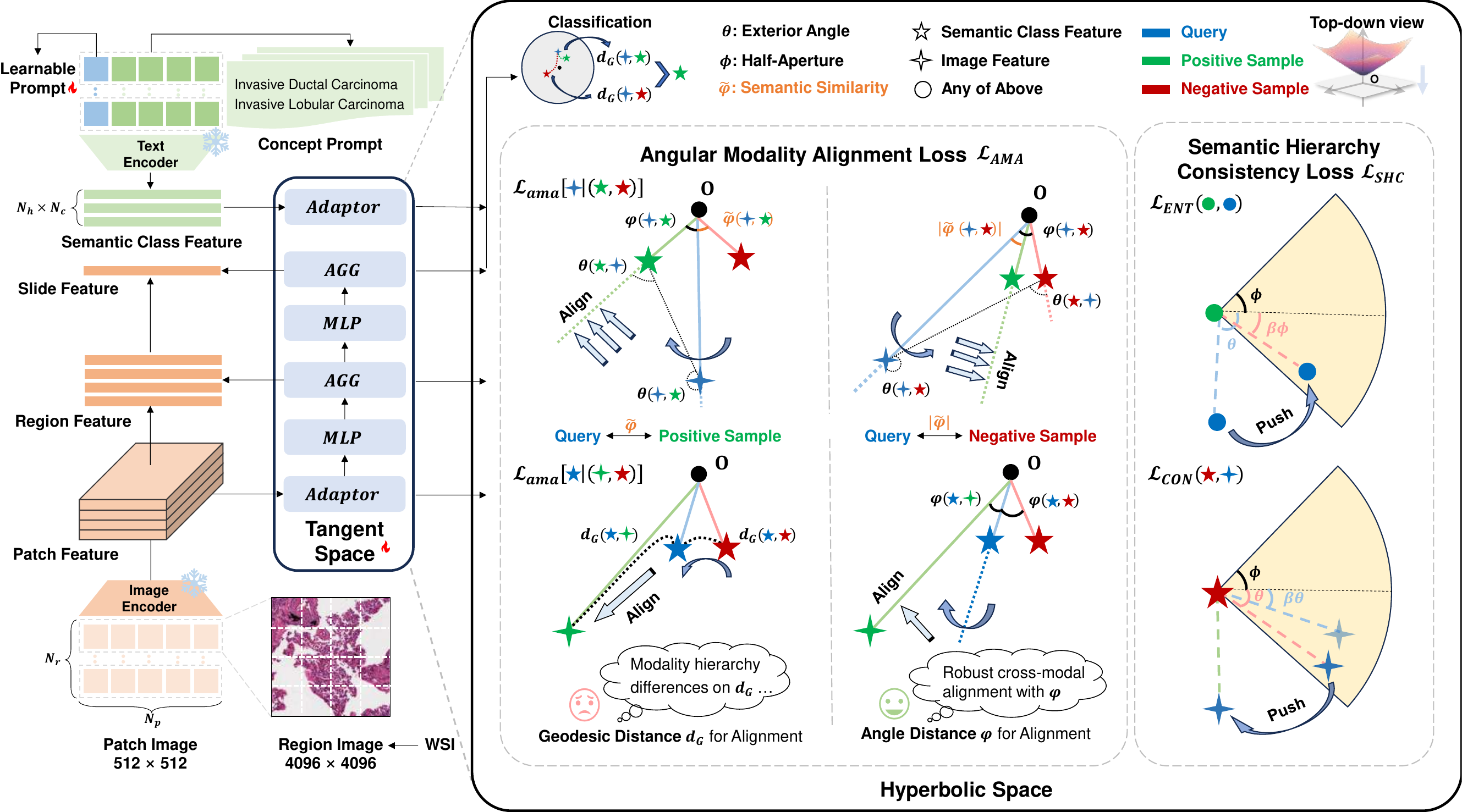}
\caption{Overview of our proposed HyperPath framework. The WSI images are hierarchically aggregated, simultaneously optimized in hyperbolic space. Guided by semantic class feature extracted from textual concepts, we utilize Angular Modality Alignment Loss and Semantic Hierarchy Consistency Loss to learn semantic hierarchies in WSIs.}
\label{fig:framework}
\end{figure}

\subsection{Overview of HyperPath}
As shown in Fig.~\ref{fig:framework}, we transfer the latent knowledge from pathology Vision-Language Models (VLMs) and adapt it to hyperbolic space, and hierarchically aggregate features. 

Specifically, the text encoder $ F_T $ extracts semantic features $ f^T_{c,h}$ through $F_T([\tilde{p}_{c,h}, p_c]) $ by combining concepts $ p_c $ with learnable prompts $ \tilde{p}_{c,h} $. These features are adapted to tangent space via $ Adaptor_T $ (a trainable two-layer MLP), producing $ \tilde{f}^T_{c,h} $, and further transformed into hyperbolic embeddings $ \mathbf{z}^T_{c,h} $ using $ \mathcal{T}_{\mathbb{E} \rightarrow \mathbb{H}} $.
For images, WSIs are divided into $ N_r $ regions ($ 4096 \times 4096 $), each split into $ N_p $ patches ($ 512 \times 512 $). Features $ f^I_p $ from $ F_I $ are mapped to $ \tilde{f}^I_p \in \mathbb{E}_{\mathbb{T}_{\mathbf{o}}} $ via $ Adaptor_I $ and embedded into $ \mathbf{z}^I_p \in \mathbb{H}_{\rho}^k  $, resulting in patch-level features of shape $ N_r \times N_p \times D $.

Then, we design an aggregator $AGG$ to integrate features $f^I_{h^{\prime}} \in \mathbb{R}^{N_h \times N_{h^{\prime}} \times D}$ from the \textbf{hierarchical subordinate level} $\textbf{h}^{\prime}$ (patch/region) to generate representations $f^I_h \in \mathbb{R}^{N_h \times 1 \times D}$ for the current level $h$ (region/slide). The process uses learnable weights $W_1 \in \mathbb{R}^{D/4 \times D}$, $W_2 \in \mathbb{R}^{D/4 \times 1}$, and is defined as:
\begin{equation}
f^I_h = \sum_{m=1}^{N_{h^\prime}} \frac{\exp\left(W^\top_2 \tanh(W_1 f^{I^\top}_{h^\prime,m})\right)}{\sum_{n=1}^{N_{h^\prime}} \exp\left(W^\top_2 \tanh(W_1 f^{I^\top}_{h^\prime,n})\right)} f^I_{h^\prime,m}.
\label{eq:eq2}
\end{equation}

After that, the aggregated features are subsequently mapped to hyperbolic space and optimized by the following Angular Modality Alignment Loss ($\mathcal{L}_{AMA}$) and Semantic Hierarchy Consistency Loss ($\mathcal{L}_{SHC}$). This process generates a more informative slide-level representation for the final prediction.

\subsection{Angular Modality Alignment Loss}
Aligning hierarchical visual and textual embeddings in hyperbolic space is crucial for cross-modal alignment, often achieved via contrastive learning methods like InfoNCE~\cite{infonce}. However, existing methods using geodesic distances~\cite{meru} struggle with modality differences.
Textual embeddings, which are more general, reside closer to the origin and entail the broad scope of concepts, while visual embeddings grow farther as granularity increases (e.g., slides$\rightarrow$regions$\rightarrow$patches).
This leads to a mismatch: intra-modal geodesic distances differ in scale from inter-modal ones, disrupting proper alignment.

To address this, we propose Angular Modality Alignment Loss $\mathcal{L}_{ama}$, which leverages angular distance instead of geodesic distance. This provides a softer way to measure semantic similarity in hierarchical structures, enabling robust cross-modal alignment despite hierarchical differences across modalities.
To be specific, we define \textbf{exterior angle} $\theta(\mathbf{u},\mathbf{v})$ as:
\begin{equation}
\theta(\mathbf{u},\mathbf{v})=\pi-\angle\mathbf{Ouv}=\cos^{-1}\left(\frac{\mathbf{v}_t+\mathbf{u}_t\rho\langle \mathbf{u}, \mathbf{v} \rangle_\mathbb{H}}{\|\mathbf{u}_s\|_\mathbb{E}\sqrt{(\rho\langle \mathbf{u}, \mathbf{v} \rangle_\mathbb{H})^2-1}}\right).
\label{eq:eq4}
\end{equation}
%
Then, this exterior angle is used to compute the \textbf{angle distance} $\varphi(\mathbf{u},\mathbf{v})=\theta(\mathbf{u},\mathbf{v})+\theta(\mathbf{v},\mathbf{u})-\pi$. 
After that, we define the \textbf{semantic similarity} as $\tilde{\varphi}(\mathbf{u},\mathbf{v}) = \varphi(\mathbf{v^+},\mathbf{v^-}) - \varphi(\mathbf{u},\mathbf{v})$, where $\mathbf{v}^+$ and $\mathbf{v}^-$ are positive and negative hyperbolic embeddings for query $\mathbf{u}$. 
Through the defined semantic similarity, we aim to push the query away from negative samples while pulling it closer to positive ones, as shown in Fig.~\ref{fig:framework}. To minimize angular distance for features of the same category, the alignment loss is formulated as:
\begin{equation}
\mathcal{L}_{ama}[\mathbf{u}|(\mathbf{v}^+,\mathbf{v}^-)] = -\log \frac{\exp({\tilde{\varphi}(\mathbf{u},\mathbf{v}^+)/\tau})}{\exp({\tilde{\varphi}(\mathbf{u},\mathbf{v}^+)/\tau}) + \sum_{\mathbf{v}^-} \exp({|\tilde{\varphi}(\mathbf{u},\mathbf{v}^-)|/\tau})},
\label{eq:eq5}
\end{equation}
where $\tau$ is the temperature. The absolute value of similarity between the query and negative samples penalizes cases where the query deviates from both negative and positive samples (Fig.~\ref{fig:framework}), avoiding suboptimal convergence. 

At each hierarchical level $h$, we apply a bidirectional alignment loss to visual and textual embeddings to mitigate bias. In the absence of specific labels for patches and regions, cosine similarity is computed using raw visual features and class semantic features extracted via CONCH. The loss is applied to the top-$K$ most similar to their corresponding class semantics.
Finally, the loss is $\mathcal{L}_{AMA}=\sum_h\left(\mathcal{L}_{ama}[\mathbf{z}^I_{h}|(\mathbf{z}^T_{c^+,h},\mathbf{z}^T_{c^-,h})]+\mathcal{L}_{ama}[\mathbf{z}^T_{c^+,h}|(\mathbf{z}^I_{h},\mathbf{z}^T_{c^-,h})]\right)$.

\subsection{Semantic Hierarchy Consistency Loss}
Beyond cross-modal alignment, capturing hierarchical semantics within and across modalities is crucial. Proper modeling ensures embeddings reflect structural dependencies and fine-grained details for coherent, interpretable representations.
To achieve this, we introduce entailment cones in hyperbolic space to model partial order relationships and reinforce hierarchical consistency. The \textbf{half-aperture} is defined as $\phi(\mathbf{u}) = \sin^{-1}\left(2\alpha/({\sqrt{\rho} \|\mathbf{u}_s\|_\mathbb{E}})\right)$, with $\alpha = 0.1$ to set boundary conditions near the origin~\cite{entailmentcone}. Based on the definition, general concepts reside closer to the origin with wider apertures, while specific concepts are farther away with narrower apertures, reflecting their hierarchy in hyperbolic space.

To maintain semantic hierarchy consistency using entailment cones, we manage both intra-modal and inter-modal relations by explicitly addressing entailment and contradiction. For semantic entailment, $\mathbf{v}$ is entailed if it lies within the entailment cone of its hierarchical superordinate $\mathbf{u}$. For semantic contradiction where $\mathbf{u}$ shouldn't entail $\mathbf{v}$, we ensure $\mathbf{v}$ remains distant from the entailment cone boundary of $\mathbf{u}$, maintaining a clear separation between conflicting semantics and strengthening hierarchical consistency. These losses are formulated as:
\begin{equation}
\left\{
\begin{aligned}
    \mathcal{L}_{ent}(\mathbf{u},\mathbf{v})&=\exp\left(\theta\left(\mathbf{u},\mathbf{v}\right)/\phi\left(\mathbf{u}\right)-1\right)\cdot \max\left(\theta\left(\mathbf{u},\mathbf{v}\right)-\beta_{ent}\cdot\phi\left(\mathbf{u}\right),0\right) \\
    \mathcal{L}_{con}(\mathbf{u},\mathbf{v})&=\exp\left(\phi\left(\mathbf{u}\right)/\theta\left(\mathbf{u},\mathbf{v}\right)-1\right)\cdot \max\left(\phi\left(\mathbf{u}\right)-\beta_{con}\cdot\theta\left(\mathbf{u},\mathbf{v}\right),0\right),
\end{aligned}
\right.
\end{equation}
where the exponential function is employed to scale the penalty, $\beta$ controls the margin, facilitating effective hierarchical semantic distinction. The final losses are defined as $\mathcal{L}_{ENT} = \sum_h \left( \mathcal{L}_{ent}(\mathbf{z}^I_{h}, \mathbf{z}^I_{h^\prime}) + \mathcal{L}_{ent}(\mathbf{z}^T_{c,h}, \mathbf{z}^T_{c,h^\prime}) + \mathcal{L}_{ent}(\mathbf{z}^T_{c^+,h}, \mathbf{z}^I_{h}) \right)$ and $\mathcal{L}_{CON} = \sum_h \left( \mathcal{L}_{con}(\mathbf{z}^T_{c^-,h}, \mathbf{z}^I_{h}) \right)$, which consist of intra- or inter-modal entailments and contradictions across hierarchical levels. Consequently, the semantic hierarchy consistency loss is given by $\mathcal{L}_{SHC} = \mathcal{L}_{ENT} + \mathcal{L}_{CON}$.

\subsection{Slide-level Prediction}

Slide-level classification is performed by leveraging semantic hierarchies in hyperbolic space, removing the dependency on additional linear classifiers. This is achieved using \textbf{geodesic} $ d_G(\mathbf{u}, \mathbf{v}) = \sqrt{1/\rho} \cosh^{-1}(-\rho \langle \mathbf{u}, \mathbf{v} \rangle_\mathbb{H}) $, which measures distances between hyperbolic embeddings as curves. Let $d_G(\mathbf{z}^I_{s}, \mathbf{z}^T_{c_i,s})$ denote the geodesic distance between slide representation $\mathbf{z}^I_{s}$ and class-specific semantics $\mathbf{z}^T_{c_i,s}$ in hyperbolic space. The classification loss $\mathcal{L}_{CLS}$ is defined as:
\begin{equation}
\mathcal{L}_{CLS} = -\sum_{i=1}^{N_C} y_i \log\left(\frac{\exp(-d_G(\mathbf{z}^I_{s}, \mathbf{z}^T_{c_i,s}))}{\sum_{j=1}^{N_C} \exp(-d_G(\mathbf{z}^I_{s}, \mathbf{z}^T_{c_j,s}))}\right).
\end{equation}

In summary, the overall loss is expressed as
$\mathcal{L} = \mathcal{L}_{CLS} + \lambda_{a} \mathcal{L}_{AMA} + \lambda_{s} \mathcal{L}_{SHC}$, where $\lambda_a$ and $\lambda_s$ balance the contributions of losses.

\section{Experiment}

\subsection{Experimental Settings}
\textbf{Datasets and Evaluation Metrics.}
We evaluated HyperPath's performance on four TCGA~\cite{tcga} tasks: breast cancer (BRCA) and non-small cell lung cancer (NSCLC) subtyping, HER2~\cite{her2} status prediction for breast cancer (BRCA HER2), and EGFR~\cite{egfr} mutation prediction for lung adenocarcinoma (LUAD EGFR). A nested splitting strategy was used: $N_{outer}$ outer folds were generated based on Tissue Source Site codes, with one fold as in-domain (IND) and the rest as out-of-domain (OOD) from entirely different sites. Within each IND fold, $N_{inner}$ inner Monte Carlo cross-validation splits were performed, resulting in $N_{outer} \times N_{inner}$ total folds. Performance was measured by mean and standard deviation of \textbf{AUC} ($\mathcal{A}$) and \textbf{F1 score} ($\mathcal{F}$) across all folds. We set $N_{outer}=3$ and $N_{inner}=5$ to ensure sufficient training samples due to limited EGFR data sites. For other tasks, $N_{outer}=5$ and $N_{inner}=3$ were used for a robust and comprehensive evaluation.

\para{Implementation Details.}
Experiments were run on a single NVIDIA RTX 3090 GPU for 20 epochs using Adam optimizer~\cite{adam} ($lr = 2 \times 10^{-4}$). Key hyperparameters were set as follows: $\tau = 0.05$, $\beta = 0.8$, $\lambda_a = 1$, and $\lambda_s = 10$. 

\begin{table}[t!]
    \centering
    \caption{Performance evaluation for different tasks. (Best: \textbf{Bold}, Second: \underline{Underlined})}
    \label{tab:performance_metrics}
    \renewcommand{\arraystretch}{0.5}
    \setlength{\heavyrulewidth}{0.12em} 
    \setlength{\lightrulewidth}{0.10em}
    \fontsize{9pt}{8pt}\selectfont 
    \begin{tabularx}{\textwidth}{l *{8}{>{\centering\arraybackslash}X}}
        \toprule
        \multirow{2}{*}{Method} & \multicolumn{4}{c}{BRCA TYPE} & \multicolumn{4}{c}{NSCLC TYPE} \\
        \cmidrule(lr){2-5} \cmidrule(l){6-9}
         & $\mathcal{A}_{OOD}$ & $\mathcal{F}_{OOD}$ & $\mathcal{A}_{IND}$ & $\mathcal{F}_{IND}$ & $\mathcal{A}_{OOD}$ & $\mathcal{F}_{OOD}$ & $\mathcal{A}_{IND}$ & $\mathcal{F}_{IND}$ \\
        \midrule
        \makecell[lc]{ABMIL~\cite{abmil}}& \meanstd{0.898}{0.040} & \meanstd{0.653}{0.070} & \meanstd{0.922}{0.058} & \meanstd{0.690}{0.151} & \meanstd{0.940}{0.030} & \meanstd{0.866}{0.027} & \meanstd{0.978}{0.014} & \meanstd{0.917}{0.032} \\
        
        \makecell[lc]{CLAM-SB~\cite{clam}} & \meanstd{0.911}{0.030} & \meanstd{\underline{0.695}}{0.061} & \meanstd{\textbf{0.934}}{0.041} & \meanstd{0.705}{0.104} & \meanstd{0.947}{0.017} & \meanstd{\underline{0.874}}{0.023} & \meanstd{0.977}{0.014} & \meanstd{0.913}{0.017} \\

        \makecell[lc]{TransMIL~\cite{transmil}} & \meanstd{0.914}{0.027}& \meanstd{0.667}{0.046}& \meanstd{0.923}{0.047}& \meanstd{0.706}{0.116}& \meanstd{0.938}{0.019} & \meanstd{0.857}{0.026} & \meanstd{0.979}{0.020} & \meanstd{0.926}{0.038} \\
        
         \makecell[lc]{DTFD-MIL~\cite{dtfdmil}} & \meanstd{0.904}{0.019} & \meanstd{0.634}{0.031} & \meanstd{0.916}{0.056} & \meanstd{0.676}{0.130} & \meanstd{0.947}{0.025} & \meanstd{0.871}{0.031} & \meanstd{0.977}{0.016} & \meanstd{0.907}{0.028} \\
        
        \makecell[lc]{ACMIL~\cite{acmil}} & \meanstd{0.897}{0.028} & \meanstd{0.645}{0.065} & \meanstd{\underline{0.931}}{0.044} & \meanstd{\underline{0.718}}{0.098} & \meanstd{0.944}{0.019} & \meanstd{0.863}{0.025} & \meanstd{0.977}{0.014} & \meanstd{0.925}{0.023} \\
        \hdashline
         \makecell[lc]{HIPT~\cite{hipt}}   & \meanstd{0.914}{0.022} & \meanstd{0.661}{0.033} & \meanstd{0.918}{0.049} & \meanstd{0.707}{0.089} & \meanstd{\underline{0.948}}{0.017} & \meanstd{0.867}{0.025} & \meanstd{0.981}{0.016} & \meanstd{0.916}{0.031} \\
         
        \makecell[lc]{HIT~\cite{conslide}}   & \meanstd{\underline{0.922}}{0.015} & \meanstd{0.679}{0.059} & \meanstd{0.920}{0.055} & \meanstd{0.705}{0.136} & \meanstd{0.938}{0.021} & \meanstd{0.864}{0.030} & \meanstd{\underline{0.982}}{0.020} & \meanstd{\underline{0.933}}{0.034} \\
       
        \midrule
        
        \textbf{HyperPath} & \meanstd{\textbf{0.933}}{0.017} & \meanstd{\textbf{0.696}}{0.060} & \meanstd{\textbf{0.934}}{0.046} & \meanstd{\textbf{0.750}}{0.086} & \meanstd{\textbf{0.957}}{0.014} & \meanstd{\textbf{0.883}}{0.017} & \meanstd{\textbf{0.984}}{0.011} & \meanstd{\textbf{0.938}}{0.026} \\
        \midrule
        \midrule
        \multirow{2}{*}{Method} & \multicolumn{4}{c}{BRCA HER2} & \multicolumn{4}{c}{LUAD EGFR} \\
        \cmidrule(lr){2-5} \cmidrule(l){6-9}
         & $\mathcal{A}_{OOD}$ & $\mathcal{F}_{OOD}$ & $\mathcal{A}_{IND}$ & $\mathcal{F}_{IND}$ & $\mathcal{A}_{OOD}$ & $\mathcal{F}_{OOD}$ & $\mathcal{A}_{IND}$ & $\mathcal{F}_{IND}$ \\
        \midrule
        \makecell[lc]{ABMIL~\cite{abmil}}
        & \meanstd{0.660}{0.049} 
        & \meanstd{0.184}{0.070} 
        & \meanstd{0.681}{0.157} 
        & \meanstd{0.210}{0.200} 
        & \meanstd{0.611}{0.039} 
        & \meanstd{0.308}{0.088} 
        & \meanstd{0.595}{0.107} 
        & \meanstd{0.315}{0.107} \\

        \makecell[lc]{CLAM-SB~\cite{clam}}
        & \meanstd{0.677}{0.062} 
        & \meanstd{0.194}{0.051} 
        & \meanstd{0.688}{0.157} 
        & \meanstd{0.229}{0.192} 
        & \meanstd{0.612}{0.045} 
        & \meanstd{0.323}{0.091} 
        & \meanstd{0.628}{0.130} 
        & \meanstd{0.320}{0.187} \\

        \makecell[lc]{TransMIL~\cite{transmil}}
        & \meanstd{0.734}{0.049} 
        & \meanstd{0.192}{0.085} 
        & \meanstd{0.700}{0.160} 
        & \meanstd{0.152}{0.163} 
        & \meanstd{0.626}{0.038} 
        & \meanstd{0.294}{0.104} 
        & \meanstd{0.619}{0.110} 
        & \meanstd{0.306}{0.164} \\

        \makecell[lc]{DTFD-MIL~\cite{dtfdmil}}
        & \meanstd{0.712}{0.060} 
        & \meanstd{\underline{0.232}}{0.052} 
        & \meanstd{0.719}{0.138} 
        & \meanstd{0.218}{0.136} 
        & \meanstd{0.573}{0.057} 
        & \meanstd{0.290}{0.123} 
        & \meanstd{0.626}{0.113} 
        & \meanstd{0.310}{0.146} \\

        \makecell[lc]{ACMIL~\cite{acmil}} 
        & \meanstd{0.716}{0.043} 
        & \meanstd{0.204}{0.057} 
        & \meanstd{0.709}{0.160} 
        & \meanstd{0.226}{0.163} 
        & \meanstd{0.612}{0.045} 
        & \meanstd{0.322}{0.117} 
        & \meanstd{0.617}{0.132} 
        & \meanstd{0.328}{0.173} \\
        
        \hdashline

        \makecell[lc]{HIPT~\cite{hipt}} 
        & \meanstd{0.732}{0.055} 
        & \meanstd{0.229}{0.060} 
        & \meanstd{0.720}{0.145} 
        & \meanstd{\underline{0.238}}{0.168} 
        & \meanstd{0.599}{0.036} 
        & \meanstd{\underline{0.343}}{0.101} 
        & \meanstd{0.630}{0.127} 
        & \meanstd{\textbf{0.358}}{0.149} \\
        
        \makecell[lc]{HIT~\cite{conslide}} 
        & \meanstd{\underline{0.740}}{0.057} 
        & \meanstd{0.075}{0.079} 
        & \meanstd{\textbf{0.740}}{0.144} 
        & \meanstd{0.093}{0.178} 
        & \meanstd{\textbf{0.638}}{0.037} 
        & \meanstd{0.237}{0.079} 
        & \meanstd{\textbf{0.647}}{0.111} 
        & \meanstd{0.256}{0.224} \\
        
        \midrule
        \textbf{HyperPath} 
        & \meanstd{\textbf{0.752}}{0.050} 
        & \meanstd{\textbf{0.260}}{0.086} 
        & \meanstd{\underline{0.732}}{0.157} 
        & \meanstd{\textbf{0.274}}{0.180} 
        & \meanstd{\underline{0.637}}{0.044} 
        & \meanstd{\textbf{0.378}}{0.093} 
        & \meanstd{\underline{0.638}}{0.107} 
        & \meanstd{\underline{0.343}}{0.120} \\
        
        \bottomrule
    \end{tabularx}
\label{tab:main_result}
\end{table}
\subsection{Experimental Results}
\para{Comparison Results.}  
As shown in Table~\ref{tab:main_result}, HyperPath was compared with state-of-the-art WSI analysis methods, including non-hierarchical (ABMIL~\cite{abmil}, CLAM-SB~\cite{clam}, TransMIL~\cite{transmil}, DTFD-MIL~\cite{dtfdmil}, ACMIL~\cite{acmil}) and hierarchical approaches (HIT~\cite{conslide}, HIPT~\cite{hipt}). All methods used CONCH~\cite{conch} for feature extraction to ensure fair comparison. 
Notably, HyperPath achieves significant gains in both AUC and F1 Score across all tasks. In the OOD setting, it outperforms others by 1.9$\%$–9.2$\%$ in AUC and 2.6$\%$–8.8$\%$ in F1 Score (excluding HIT's outlier). Similarly, in the IND setting, it shows improvements of 0.7$\%$–5.1$\%$ in AUC and 3.1$\%$–12.2$\%$ in F1 Score. This consistent performance highlights its robustness with minimal variation between IND and OOD scenarios, except for a slight drop in $\mathcal{F}_{IND}$ on LUAD EGFR, likely due to small sample size and class imbalance. 
While HIT achieves high AUC in BRCA HER2 and LUAD EGFR, its low F1 score indicates a bias toward certain classes, reflecting overfitting and instability. In contrast, HyperPath delivers balanced and reliable results, excelling in both metrics and demonstrating its superiority across diverse tasks.

\para{Ablation Analysis.}  
We conduct ablation studies to evaluate the effectiveness of $\mathcal{L}_{AMA}$ and $\mathcal{L}_{SHC}$ as shown in Tab.~\ref{tab:ablation}. Using $\mathcal{L}_{AMA}$ alone improves performance by aligning hyperbolic visual features with class semantics, while $\mathcal{L}_{SHC}$ alone degrades performance due to neglecting visual-textual alignment, leading to scattered feature distributions. Combining both losses achieves optimal results, as $\mathcal{L}_{AMA}$ ensures precise visual-semantic alignment and $\mathcal{L}_{SHC}$ enhances semantic hierarchies, promoting intra-semantic alignment, inter-semantic separation, and ultimately boosting classification performance through jointly learned multi-modal hyperbolic features with the consistent semantic hierarchy.

\begin{figure}[t!]
\centering
\includegraphics[width=0.92\textwidth]{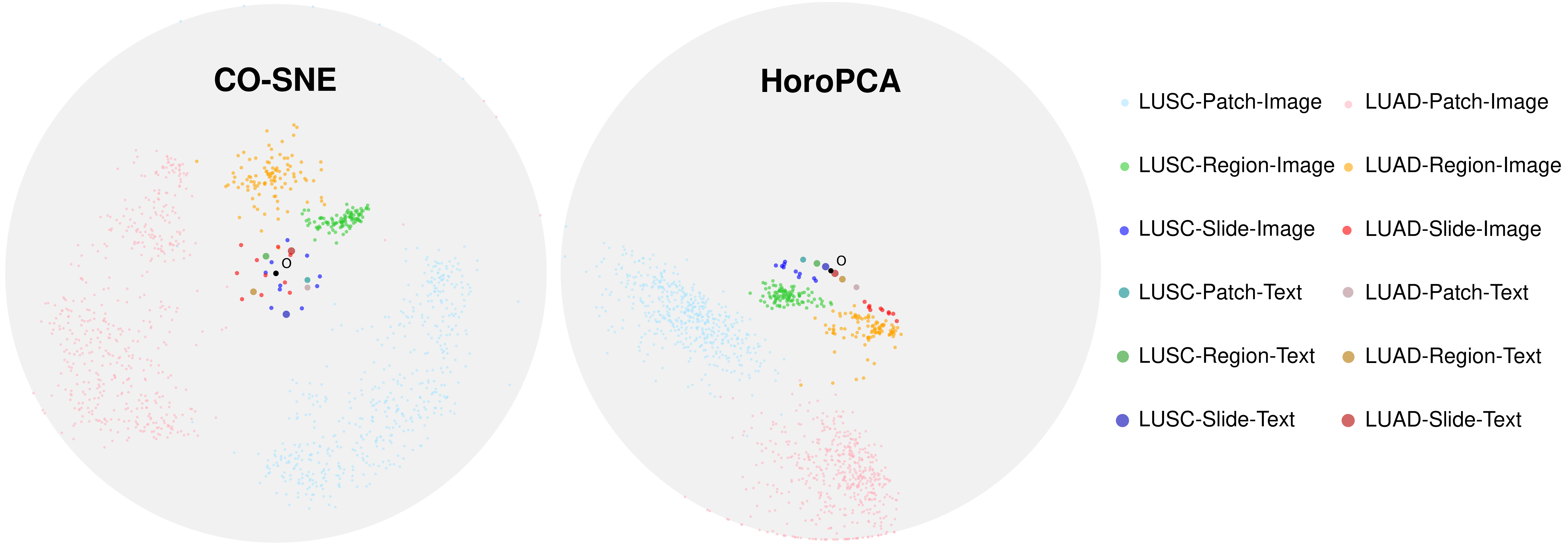}
\caption{The visualization of hyperbolic embeddings from different hierarchical levels. It is observed that the embeddings are well-structured in hyperbolic space.}
\label{fig:vis}
\end{figure}

\para{Hyperbolic Embedding Visualization.}
Fig.~\ref{fig:vis} visualizes hyperbolic embeddings using dimensionality reduction methods CO-SNE~\cite{cosne} and HoroPCA~\cite{horopca} designed for hyperbolic space. Using NSCLC subtyping task as an example, we display features across modalities, categories, and hierarchical levels. The visualization reveals a clear hierarchical structure: class semantic features cluster near the origin, surrounded by slide-, region-, and patch-level features in order. Distinct category boundaries confirm that our method effectively aligns and distributes multimodal features in hyperbolic space. Such hierarchical representations can further enhance WSI classification performance.

\begin{table}[t!]
    \centering
    \caption{Ablation study of HyperPath.}
    \renewcommand{\arraystretch}{0.5} 
    \fontsize{9pt}{8pt}\selectfont  
    \begin{tabularx}{\textwidth}{cc*{8}{>{\centering\arraybackslash}X}}
        \toprule
        \multicolumn{2}{c}{HyperPath} & \multicolumn{4}{c}{BRCA} & \multicolumn{4}{c}{NSCLC} \\
         \cmidrule(lr){3-6} \cmidrule(l){7-10} 
        $\mathcal{L}_{{AMA}}$ & $\mathcal{L}_{{SHC}}$ & $\mathcal{A}_{OOD}$ & $\mathcal{F}_{OOD}$ & $\mathcal{A}_{IND}$ & $\mathcal{F}_{IND}$ & $\mathcal{A}_{OOD}$ & $\mathcal{F}_{OOD}$ & $\mathcal{A}_{IND}$ & $\mathcal{F}_{IND}$ \\
        \midrule
        & & \meanstd{0.864}{0.032} & \meanstd{0.532}{0.151} & \meanstd{0.893}{0.055} & \meanstd{0.564}{0.199} & \meanstd{0.849}{0.120} & \meanstd{0.814}{0.073} & \meanstd{0.914}{0.106} & \meanstd{0.887}{0.081} \\
        \makecell[c]{\checkmark} & & \meanstd{0.925}{0.016} & \meanstd{0.634}{0.107} & \meanstd{0.928}{0.046} & \meanstd{0.656}{0.150} & \meanstd{0.947}{0.021} & \meanstd{0.832}{0.114} & \meanstd{0.975}{0.024} & \meanstd{0.889}{0.121} \\
        & \makecell[c]{\checkmark} & \meanstd{0.539}{0.105} & \meanstd{0.156}{0.141} & \meanstd{0.558}{0.123} & \meanstd{0.153}{0.139} & \meanstd{0.499}{0.099} & \meanstd{0.302}{0.326} & \meanstd{0.507}{0.156} & \meanstd{0.331}{0.331} \\
        \makecell[c]{\checkmark} & \makecell[c]{\checkmark} & \meanstd{\textbf{0.933}}{0.017} & \meanstd{\textbf{0.696}}{0.060} & \meanstd{\textbf{0.934}}{0.046} & \meanstd{\textbf{0.750}}{0.086} & \meanstd{\textbf{0.957}}{0.014} & \meanstd{\textbf{0.883}}{0.017} & \meanstd{\textbf{0.984}}{0.011} & \meanstd{\textbf{0.938}}{0.026} \\
        \bottomrule
    \end{tabularx}
\label{tab:ablation}
\end{table}

\section{Conclusion}
In this paper, we introduce HyperPath, a novel approach that leverages hyperbolic space to learn hierarchical representations of WSIs. HyperPath aggregates patch-level features from a pathology vision-language foundation model into region- and slide-level representations. Through angular modality alignment loss, semantically similar features are brought closer in hyperbolic space, while a semantic hierarchy consistency loss enhances inter- and intra-modality relationships, yielding meaningful hierarchies. Experiments demonstrate that HyperPath surpasses existing methods across multiple tasks, showing the potential of hyperbolic space as a powerful alternative for modeling complex WSIs.

\begin{credits}
\subsubsection{\ackname} This work was supported in part by the Research Grants Council of Hong Kong (27206123, C5055-24G, and T45-401/22-N), the Hong Kong Innovation and Technology Fund (ITS/274/22, and GHP/318/22GD), the National Natural Science Foundation of China (No. 62201483), and Guangdong Natural Science Fund (No. 2024A1515011875).

\subsubsection{\discintname}
The authors have no competing interests to declare that
are relevant to the content of this article.
\end{credits}

%
%
%
\bibliographystyle{splncs04}
\bibliography{ref}

\end{document}